# A Paradigm for Potential Model Performance Improvement in Classification and Regression Problems. A Proof of Concept

Francisco Javier Lobo-Cabrera

A methodology that seeks to enhance model prediction performance is presented. The method involves generating multiple auxiliary models that capture relationships between attributes as a function of each other. Such information serves to generate additional informative columns in the dataset that can potentially enhance target prediction. A proof of case and related code is provided.

## Introduction

Binary classification, multilabel classification, and regression prediction constitute fundamental paradigms in machine learning, addressing distinct types of predictive modeling tasks. Binary classification involves categorizing instances into one of two classes, typically denoted as positive and negative [1][2][3]. This modeling framework is particularly applicable to scenarios where outcomes are binary in nature, as observed in domains such as spam detection and medical diagnosis. In multilabel classification, the scope extends to situations where instances can be associated with multiple classes simultaneously, a common occurrence in applications like image tagging and document categorization [1][4]. Conversely, regression prediction is concerned with forecasting continuous outcomes, aiming to predict numeric values [3]. This modeling approach is prevalent in scenarios where the target variable is a real number, as exemplified in the prediction of housing prices based on diverse features [3][5]. While binary and multilabel classification focus on assigning categorical labels, regression models emphasize the estimation of a numeric response. The existence of these varied approaches serves to cater to the diverse array of challenges encountered in machine learning applications, providing practitioners with versatile tools to address a multitude of real-world problems.

While the evolution of classification and regression methods has been marked by advancements, persistent challenges pose hurdles to achieving optimal performance. Imbalanced class distribution is a common challenge, where one class significantly outnumbers the other [6]. In scenarios where the dataset is skewed, models may exhibit a bias towards the majority class, resulting in diminished predictive accuracy for the minority class [6]. Noisy data further compounds the challenges in target prediction. Real-world datasets are often fraught with errors, outliers, or inaccuracies, impacting the model's ability to discern meaningful patterns [7]. The presence of noisy data can lead to suboptimal generalization and reduced reliability of the model [7]. Another challenge arises from the complexity of decision boundaries in certain classification problems, which has prompted the development of various algorithms to address these situations [8][9].

In summary, in spite of constant evolution of predictive algorithms, challenges persist for some datasets in achieving robust and accurate models. Imbalanced class distribution, noisy data, and complex decision boundaries demand ongoing research into innovative methodologies and a nuanced understanding of the underlying intricacies. The dynamic nature of these challenges propels the field towards continuous innovation and the development of adaptive strategies for improved target prediction performance.

Here, a methodology for increasing the predictive power of a given dataset is presented, providing a proof case to demonstrate its validity. The methodology is based on the assumption that attribute variables relate to each other differently according to the value of the target. Such relations can be modeled and trained in turn using any standard ML algorithms, neural networks or other related methods. The different modeled relations among attributes as a function of the target can be exploited to acquire new auxiliary variables that possess predictive power on the target itself. In this manner, the original dataset can be augmented with additional columns that may increase, at least potentially, model performance.

**Proof of case**

In this example, the Bank Marketing dataset from the UCI Machine Learning Repository was employed [10]. The dataset pertains to the direct marketing campaigns conducted by a Portuguese banking institution. These campaigns primarily involved reaching out to clients through phone calls, determining whether they would subscribe to the bank term deposit, indicated by responses of 'yes' or 'no', which are registered in the dataset variable 'y'. The rest of the variables in the dataset are considered attribute variables from which a predictive model on 'y' can be constructed using supervised learning.

Standard machine-learning procedures were carried out in the dataset, including elimination of entries that were either duplicated or that contain non valid values and one-hot encoding of categorical variables to render a total of 63 attribute variables. The variables were also scaled to display a mean value of 0 and a standard deviation of 1. In addition, the data was sampled (random selection of 60% of the rows) to speed up subsequent calculations. The dataset was then split into two datasets (A_1 and A_0), A_1 containing the registers with target value = 1 and A_0 including those registers with target value = 0. Then, for each of the datasets created (A_1 and A_0) a series of models were constructed that predicted each of the attribute variables as a function of the rest of the attribute variables. Since after the one-hot encoding there were 63 attributes and in this case being a binary classification two dataframes were created (A_1 and A_0), the number of created models equals 63x2 = 126. All of the created models and the associated goodness of fit ($R^2$) were saved.

At this point, the created models were applied to the entire dataset (A) to predict the value of each of its attribute variables. These predictions were compared to the actual attribute variables and the difference between them in the form of RMSE, i.e by squaring the difference and calculating the square root, was computed. Such computed differences between predicted attribute values according to the created models and the actual values were stored as a new columns. In particular, for each attribute, the difference between the model that predicts it from dataset A_0 and the actual value was stored as a new column, and the difference between the model that predicts it from dataset A_1 and the actual value was stored as another new column. In this manner, for each original attribute column two new columns were created.

Also, it must be mentioned that the differences in this example were weighed according to the dispersion of the model to consider the fitness of the model (calculated from R²), to penalize more heavily the difference when the model predicting the attribute is highly accurate.

In the provided code (see Section Appendix I. Example code), each of the new columns generated are named after the original attribute column adding '_1' or '_0' as a suffix depending on the dataset (A_1 or A_0) from which the model in question was created. In this way, each row of dataset A will now contain i) the normal attribute variables, ii) the target variable and iii) the variables created by comparing the normal attributes to the predicted attributes according to the models generated from A_0 and A_1. It is important to note that the normal attribute variables and the new columns added may have high correlation, but not necessarily 100%, so that the added columns are potentially providing more information to the dataset.

In this manner, the newly transformed dataset A now contains more informative columns. From this updated dataset, a standard predictive model can be constructed to predict the target. Since more predictive variables have become available, it is expected accordingly that model performance may be potentially higher, which is in fact achieved in this example as can be seen after execution of the code, reaching a new F1-score value of 0.82376. Nevertheless, if the ratio number of rows / number of columns were too low (usually less than 10) the new model performance may be hindered. For this reason, it is advisable that the original number of rows in the dataframe is high enough to accommodate the new columns.

Importantly, the process may be repeated so that more new columns can be added to the dataset. Nevertheless, as stated before, the generation of more columns means that the ratio number of rows / number of columns decreases, so that the cycle may be repeated only for a certain number of rounds.

In a second example, an analogous methodology is employed instead for a regression task. For this purpose, the same dataset as described for binary classification is employed, although in this case the target variable chosen is 'age'. Also, only one predictive model for each attribute is generated, instead of the two (one for each target value) developed in the binary classification example. The results for the normalized target variable shows that the augmented dataset renders in this case slightly inferior model performance (RMSE = 0.00402) to that generated from the original dataset (RMSE = 0.00397). So, in this example, the addition of the new columns has only served to add noise to the prediction. This highlights that the referred process may be useful or not depending on the particular dataset and target.

In the examples provided (see also section Appendix I. Example Code) Random Forest classifiers and regressors are employed as an illustration, but the choice of algorithm to utilize may change depending on the classification dataset and user preferences.

## Discussion

Classification and regression problems are frequently tackled generating predictive models from a supervised learning datasets. These models capture the relations between attributes to predict the target value. Nevertheless, attribute values may also have an impact on the actual value of the rest of attributes, and these relations may also be modeled. Here, a simple proof of case in binary classification achieves this and by doing so incorporates more informative columns to the dataset, effectively enhancing model effectiveness. The process may also be extended easily to multilabel classification and regression problems. However, in an example on a regression task, the added columns in fact slightly decreased model performance, highlighting that the described process may offer utility or not depending on the characteristics of the available data.

In summary, this proof of case highlights that target prediction may be enhanced by creating models that do not predict the target directly, but rather indirectly, examining how the rest of the variables interact with each other according to its value. This approach, although computationally more expensive, may be useful in datasets where interactions between attributes display distinct behavior depending on the value of the target.

## References

[1] Saraswat, P. (2022). Supervised Machine Learning Algorithm: A Review of Classification Techniques. In International Conference on Intelligent Emerging Methods of Artificial Intelligence & Cloud Computing (pp. 477–482). Springer International Publishing. https://doi.org/10.1007/978-3-030-92905-3_58

[2] Li, J. J., & Tong, X. (2020). Statistical Hypothesis Testing versus Machine Learning Binary Classification: Distinctions and Guidelines. In Patterns (Vol. 1, Issue 7, p. 100115). Elsevier BV. https://doi.org/10.1016/j.patter.2020.100115

[3] Woodman, R. J., & Mangoni, A. A. (2023). A comprehensive review of machine learning algorithms and their application in geriatric medicine: present and future. In Aging Clinical and Experimental Research (Vol. 35, Issue 11, pp. 2363–2397). Springer Science and Business Media LLC. https://doi.org/10.1007/s40520-023-02552-2

[4] Hicks, S. A., Strümke, I., Thambawita, V., Hammou, M., Riegler, M. A., Halvorsen, P., & Parasa, S. (2022). On evaluation metrics for medical applications of artificial intelligence. In Scientific Reports (Vol. 12, Issue 1). Springer Science and Business Media LLC. https://doi.org/10.1038/s41598-022-09954-8

[5] Pineda-Jaramillo J. D. (2019) A Review of Machine Learning (ML) Algorithms Used for Modeling Travel Mode Choice (Vol. 86, No. 211, 2019, pp. 32–41) Dyna. https://doi.org/10.15446/dyna

[4] Sarker, I. H. (2021). Machine Learning: Algorithms, Real-World Applications and Research Directions. In SN Computer Science (Vol. 2, Issue 3). Springer Science and Business Media LLC. https://doi.org/10.1007/s42979-021-00592-x

[5] Foryś, I. (2022). Machine learning in house price analysis: regression models versus neural networks. In Procedia Computer Science (Vol. 207, pp. 435–445). Elsevier BV. https://doi.org/10.1016/j.procs.2022.09.078

[6] Johnson, J. M., & Khoshgoftaar, T. M. (2019). Survey on deep learning with class imbalance. In Journal of Big Data (Vol. 6, Issue 1). Springer Science and Business Media LLC. https://doi.org/10.1186/s40537-019-0192-5

[7] Prati, R. C., Luengo, J., & Herrera, F. (2018). Emerging topics and challenges of learning from noisy data in nonstandard classification: a survey beyond binary class noise. In Knowledge and Information Systems (Vol. 60, Issue 1, pp. 63–97). Springer Science and Business Media LLC. https://doi.org/10.1007/s10115-018-1244-4

[8] Kumagai, A., & Iwata, T. (2017). Learning Non-Linear Dynamics of Decision Boundaries for Maintaining Classification Performance. In Proceedings of the AAAI Conference on Artificial Intelligence (Vol. 31, Issue 1). Association for the Advancement of Artificial Intelligence (AAAI). https://doi.org/10.1609/aaai.v31i1.10830

[9] Karimi, H., & Tang, J. (2020). Decision Boundary of Deep Neural Networks. In Proceedings of the 13th International Conference on Web Search and Data Mining. WSDM '20: The Thirteenth ACM International Conference on Web Search and Data Mining. ACM. https://doi.org/10.1145/3336191.3372186

[10] Moro,S., Rita,P., and Cortez,P.. (2012). Bank Marketing. UCI Machine Learning Repository. https://doi.org/10.24432/C5K306.

**Appendix I. Example Code**

Note: Future versions of the code will be optimized (calculations in matrix form) so that execution time is decreased.

a) Code for binary classification example:

Note: Although provided here, due to the formatting of the document tt is recommended rather to access the code at:

https://github.com/UPO-Sevilla-Fco-Javier-Lobo-Cabrera/Juxtaposed_Attribute_Ensemble_Networks/blob/main/juxtaposed_attribute_ensemble_networks_V3.py

```python
import numpy as np
import pandas as pd
from sklearn.model_selection import train_test_split, cross_val_predict
from sklearn.ensemble import RandomForestRegressor
from sklearn.ensemble import RandomForestClassifier
from sklearn.metrics import precision_score, recall_score, f1_score, roc_auc_score
from sklearn.preprocessing import StandardScaler
from sklearn.metrics import mean_squared_error
import zipfile
import io
import requests
import time

############################
# Download dataset from UCI* for the first time and save it locally:
print("Obtaining dataset...")
url = 'https://archive.ics.uci.edu/ml/machine-learning-databases/00222/bank-additional.zip'
response = requests.get(url)
z = zipfile.ZipFile(io.BytesIO(response.content))
# Specify the name of the CSV file to read from the ZIP file
csv_filename = 'bank-additional/bank-additional.csv'
# Read the specified CSV file into a DataFrame
with z.open(csv_filename) as file:
    data = pd.read_csv(file, sep=';')

#(*)Moro,S., Rita,P., and Cortez,P.. (2012). Bank Marketing. UCI Machine
# Learning Repository. https://doi.org/10.24432/C5K306.
############################

# Sampling of the data
data = data.sample(frac=0.6, random_state=42)

data = data.drop_duplicates()

# Handling missing values (drop rows with missing values for simplicity)
data.dropna(inplace=True)

# Encoding categorical variables using one-hot encoding
data = pd.get_dummies(data)

# Name of the target of the dataset (after one-hot encoding)
target_variable = 'y_yes'
```

```python
# Normalizing variables:
data_columns = data.columns
scaler = StandardScaler()
data = scaler.fit_transform(data)
data = pd.DataFrame(data, columns=data_columns)

# Denormalize target values (these must be 0 or 1):
def aux_denormalize_target(aux):
    threshold = min(list(data[target_variable].unique()))
    if aux > threshold:
        return 1
    else:
        return 0

data[target_variable] = data[target_variable].apply(aux_denormalize_target)

# Drop the other column obtained from one-hot encoding the target:
data = data.drop(columns=["y_no"])

# List of dictionaries
list_of_dictionaries = []

# List of dictionaries of R²
list_of_dictionaries_r_squared = []

# For each value of the target
for target_value in sorted(list(data[target_variable].unique())):
    print("\nCreating attribute models for cases when target is " + str(target_value) + "...\n#########################")

    # Generate auxiliary dataset
    dataset_aux = data[data[target_variable] == target_value]

    # Discard target in auxiliary dataset
    dataset_aux = dataset_aux.drop(columns=[target_variable])

    # Generate dictionary of ficticious targets and the models that predict them:
    dictionary_aux = {}
    # Correspondant dictionary of rmse for weighing
    dictionary_aux_r_squared = {}

    # Creation of model for each ficticious target (i.e attribute variable)
    for fict_target in dataset_aux.columns.tolist():
        print("Creating model for " + str(fict_target) + "...")

        # Train the Random Forest model and save it
        X = dataset_aux.drop(columns=[fict_target])
```

```python
        y = dataset_aux[fict_target]

        X_train, X_test, y_train, y_test = train_test_split(X, y, test_size=0.2, random_state=42)

        # Fit a regressor:
        if True:
            rf = RandomForestRegressor(n_estimators=100, random_state=42)
            rf.fit(X_train, y_train)
            dictionary_aux[fict_target] = rf

            #####
            # Computation of R² for weighing:
            predictions = rf.predict(X_test)
            y_mean = np.mean(y_test)
            # Calculate the total sum of squares
            tss = np.sum((y_test - y_mean) ** 2)
            # Calculate the residual sum of squares
            rss = np.sum((y_test - predictions) ** 2)
            # Calculate R² score
            # If tss == 0 then R² will be 1
            if (tss < 0.00001) & (tss > -0.00001):
                r_squared = 1
            else:
                r_squared = 1 - (rss / tss)

            dictionary_aux_r_squared[fict_target] = r_squared

    list_of_dictionaries.append(dictionary_aux)
    list_of_dictionaries_r_squared.append(dictionary_aux_r_squared)

list_unique_values_target = sorted(list(data[target_variable].unique()))

list_of_rows_dataframe_new = []

# For each register calculate new columns
number_of_rows = len(data)
print("Number of rows to process: " + str(number_of_rows))
time.sleep(5)

for i in range(0, number_of_rows):
    print(i)
    row = data.iloc[i]
    # Convert the row to a dataframe object of one row
    row = row.to_frame().T

    # List of sublists of rmse (this is auxiliary for later creating the new
    # columns)
    list_rmse = []

    # For each value of the target
```

```python
    for case in range(0, len(list_unique_values_target)):

        # Obtain dictionary of attribute models and dictionary of correspondant
        # R² values of those models
        dictionary_case = list_of_dictionaries[case]
        dictionary_case_r_squared = list_of_dictionaries_r_squared[case]

        sub_list_rmse = []

        for fict_target in dictionary_case:
            # Calculation of the predicted value of the attribute
            X = row.drop(columns=[target_variable, fict_target])
            y_predicted = dictionary_case[fict_target].predict(X)
            # Calculation of difference between predicted value of the attribute
            # and actual value. The difference is calculated in form of rmse
            # and then weighed according to the R² of the predictive model for
            # the attribute in question
            y_real = row[fict_target]
            mse = (y_real - y_predicted) ** 2
            rmse = np.sqrt(mse)
            rmse = float(round(rmse.iloc[0], 4))
            # Weigh according to r squared of the model
            rmse = rmse * (dictionary_case_r_squared[fict_target])**2

            # Add to sublist
            sub_list_rmse.append(rmse)

        list_rmse.append(sub_list_rmse)

    # Generate row of dataframe containing only the new columns
    # (this dataframe will be later concatenated horizontally
    # to the original dataframe, so that the resulting dataframe
    # contains both the original and the new columns):
    row_to_append = []
    # Add the columns that contain the difference between the
    # predicted attribute value and the actual one)
    for h in range(0, len(sub_list_rmse)):
        row_to_append.append(list_rmse[0][h])
        row_to_append.append(list_rmse[1][h])

    list_of_rows_dataframe_new.append(row_to_append)

# Generate dataframe that contains only the new columns
names_cols_dataframe_new = []
for u in dictionary_case.keys():
    if (u != target_variable):
        names_cols_dataframe_new.append(u + "_0")
        names_cols_dataframe_new.append(u + "_1")
dataframe_new = pd.DataFrame(list_of_rows_dataframe_new,
columns=names_cols_dataframe_new)
```

###########################################################################
# Concatenation of new columns to the original dataframe

df = dataframe_new

cols_a = data.columns.to_list()
cols_b = df.columns.to_list()

data = data.reset_index(drop=True)
df = df.reset_index(drop=True)

# Concatenate horizontally to add new columns
result_df = pd.concat([df, data], axis=1, ignore_index=True)
result_df.columns = cols_b + cols_a

# The process has generated additional columns in the dataframe (those ending with _0 or _1).
# These additional columns could enhance potentially performance.
# The whole cycle may be repeated again (sort of a new layer) generating more additional
# variables (these will contain also those now ending with _0_0, _0_1, _1_0, and 1_1).

###########################################################################

# Comparison between performance metrics of the updated dataset and
# the normal (original) dataset. For this purpose, an RF model using
# cross validation throughout all rows is employed in both cases
# with default sklearn RF hyperparameters, which imply max_depth=None
# so that the trees can potentially grow very deep.

# Obtain metrics of the updated dataset
features = result_df.drop(target_variable, axis=1)
target = result_df[target_variable]
# Initialize RandomForestClassifier
rf_model = RandomForestClassifier(random_state=42)
# Perform k-fold cross-validation (k=5)
cv_predictions = cross_val_predict(rf_model, features, target, cv=5)
# Calculate precision, recall, and F1 scores
precision = precision_score(target, cv_predictions)
recall = recall_score(target, cv_predictions)
f1 = f1_score(target, cv_predictions)
# Print the precision, recall, and F1 scores
print(f'Precision: {precision}')
print(f'Recall: {recall}')
print(f'F1 Score: {f1}')

# Obtain metrics of the normal (i.e original) dataset
features = data.drop(target_variable, axis=1)
target = data[target_variable]
rf_normal = RandomForestClassifier(random_state=42)
# Perform k-fold cross-validation (k=5)
cv_predictions = cross_val_predict(rf_normal, features, target, cv=5)
precision = precision_score(target, cv_predictions)
recall = recall_score(target, cv_predictions)
```

```
f1 = f1_score(target, cv_predictions)
# Print the normal precision, recall, and F1 scores
print(f'Normal Precision: {precision}')
print(f'Normal Recall: {recall}')
print(f'Normal F1 Score: {f1}')
```

b) Code for regression example:

Note: Although provided here, due to the formatting of the document it is recommended rather to access the code at:

https://github.com/UPO-Sevilla-Fco-Javier-Lobo-Cabrera/Juxtaposed_Attribute_Ensemble_Networks/blob/main/juxtaposed_attribute_ensemble_networks_continuous_V1.py

```
import numpy as np
import pandas as pd
from sklearn.model_selection import train_test_split, cross_val_predict
from sklearn.ensemble import RandomForestRegressor
from sklearn.ensemble import RandomForestClassifier
from sklearn.metrics import precision_score, recall_score, f1_score, roc_auc_score
from sklearn.preprocessing import StandardScaler
from sklearn.metrics import mean_squared_error
import zipfile
import io
import requests
import time

#############################
# Download dataset from UCI* for the first time and save it locally:
print("Obtaining dataset...")
url = 'https://archive.ics.uci.edu/ml/machine-learning-databases/00222/bank-additional.zip'
response = requests.get(url)
z = zipfile.ZipFile(io.BytesIO(response.content))
# Specify the name of the CSV file to read from the ZIP file
csv_filename = 'bank-additional/bank-additional.csv'
# Read the specified CSV file into a DataFrame
with z.open(csv_filename) as file:
    data = pd.read_csv(file, sep=';')

#(*)Moro,S., Rita,P., and Cortez,P.. (2012). Bank Marketing. UCI Machine
# Learning Repository. https://doi.org/10.24432/C5K306.
#############################

# Sampling of the data
data = data.sample(frac=0.6, random_state=42)
```

```python
data = data.drop_duplicates()

# Handling missing values (drop rows with missing values for simplicity)
data.dropna(inplace=True)

# Encoding categorical variables using one-hot encoding
data = pd.get_dummies(data)

# Name of the target of the dataset (after one-hot encoding)
target_variable = 'age'

# Normalizing variables:
data_columns = data.columns
scaler = StandardScaler()
data = scaler.fit_transform(data)
data = pd.DataFrame(data, columns=data_columns)

# Denormalize target values (these must be 0 or 1):
def aux_denormalize_target(aux):
    threshold = min(list(data[target_variable].unique()))
    if aux > threshold:
        return 1
    else:
        return 0

data[target_variable] = data[target_variable].apply(aux_denormalize_target)

# List of dictionaries
list_of_dictionaries = []

# List of dictionaries of R²
list_of_dictionaries_r_squared = []

if True:
    # Generate auxiliary dataset
    dataset_aux = data.drop(columns=[target_variable])

    # Generate dictionary of ficticious targets and the models that predict them:
    dictionary_aux = {}
    # Correspondant dictionary of rmse for weighing
    dictionary_aux_r_squared = {}

    # Creation of model for each ficticious target (i.e attribute variable)
    for fict_target in dataset_aux.columns.tolist():
        print("Creating model for " + str(fict_target) + "...")

        # Train the Random Forest model and save it
        X = dataset_aux.drop(columns=[fict_target])
```

```python
            y = dataset_aux[fict_target]

            X_train, X_test, y_train, y_test = train_test_split(X, y, test_size=0.2, random_state=42)

            # Fit a regressor:
            if True:
                rf = RandomForestRegressor(n_estimators=100, random_state=42)
                rf.fit(X_train, y_train)
                dictionary_aux[fict_target] = rf

                #####
                # Computation of R² for weighing:
                predictions = rf.predict(X_test)
                y_mean = np.mean(y_test)
                # Calculate the total sum of squares
                tss = np.sum((y_test - y_mean) ** 2)
                # Calculate the residual sum of squares
                rss = np.sum((y_test - predictions) ** 2)
                # Calculate R² score
                # If tss == 0 then R² will be 1
                if (tss < 0.00001) & (tss > -0.00001):
                    r_squared = 1
                else:
                    r_squared = 1 - (rss / tss)

                dictionary_aux_r_squared[fict_target] = r_squared

        list_of_dictionaries.append(dictionary_aux)
        list_of_dictionaries_r_squared.append(dictionary_aux_r_squared)

list_unique_values_target = sorted(list(data[target_variable].unique()))

list_of_rows_dataframe_new = []

# For each register calculate new columns
number_of_rows = len(data)
print("Number of rows to process: " + str(number_of_rows))
time.sleep(5)

for i in range(0, number_of_rows):
    print(i)
    row = data.iloc[i]
    # Convert the row to a dataframe object of one row
    row = row.to_frame().T

    # List of sublists of rmse (this is auxiliary for later creating the new
    # columns)
    list_rmse = []

    if True:
```

```python
        # Obtain dictionary of attribute models and dictionary of correspondant
        # R² values of those models
        dictionary_case = list_of_dictionaries[0]
        dictionary_case_r_squared = list_of_dictionaries_r_squared[0]

        sub_list_rmse = []

        for fict_target in dictionary_case:
            # Calculation of the predicted value of the attribute
            X = row.drop(columns=[target_variable, fict_target])
            y_predicted = dictionary_case[fict_target].predict(X)
            # Calculation of difference between predicted value of the attribute
            # and actual value. The difference is calculated in form of rmse
            # and then weighed according to the R² of the predictive model for
            # the attribute in question
            y_real = row[fict_target]
            mse = (y_real - y_predicted) ** 2
            rmse = np.sqrt(mse)
            rmse = float(round(rmse.iloc[0], 4))
            # Weigh according to r squared of the model
            rmse = rmse * (dictionary_case_r_squared[fict_target])**2

            # Add to sublist
            sub_list_rmse.append(rmse)

        list_rmse.append(sub_list_rmse)

    # Generate row of dataframe containing only the new columns
    # (this dataframe will be later concatenated horizontally
    # to the original dataframe, so that the resulting dataframe
    # contains both the original and the new columns):
    row_to_append = []
    # Add the columns that contain the difference between the
    # predicted attribute value and the actual one)
    for h in range(0, len(sub_list_rmse)):
        row_to_append.append(list_rmse[0][h])

    list_of_rows_dataframe_new.append(row_to_append)

# Generate dataframe that contains only the new columns
names_cols_dataframe_new = []
for u in dictionary_case.keys():
    if (u != target_variable):
        names_cols_dataframe_new.append(u + "_new")

dataframe_new = pd.DataFrame(list_of_rows_dataframe_new,
columns=names_cols_dataframe_new)

######################################################################
# Concatenation of new columns to the original dataframe
```

```python
df = dataframe_new

cols_a = data.columns.to_list()
cols_b = df.columns.to_list()

data = data.reset_index(drop=True)
df = df.reset_index(drop=True)

# Concatenate horizontally to add new columns
result_df = pd.concat([df, data], axis=1, ignore_index=True)
result_df.columns = cols_b + cols_a

# The process has generated additional columns in the dataframe (those ending with _0 or _1).
# These additional columns could enhance potentially performance.
# The whole cycle may be repeated again (sort of a new layer) generating more additional
# variables (these will contain also those now ending with _0_0, _0_1, _1_0, and 1_1).

#######################################################################

# Comparison between performance metrics of the updated dataset and
# the normal (original) dataset. For this purpose, an RF model using
# cross validation throughout all rows is employed in both cases
# with default sklearn RF hyperparameters, which imply max_depth=None
# so that the trees can potentially grow very deep.

# Obtain metrics of the updated dataset
features = result_df.drop(target_variable, axis=1)
target = result_df[target_variable]
# Initialize RandomForestClassifier
rf_model = RandomForestRegressor(random_state=42)
# Perform k-fold cross-validation (k=5)
cv_predictions = cross_val_predict(rf_model, features, target, cv=5)
# Calculate rmse
rmse = np.sqrt(mean_squared_error(target, cv_predictions))
print("RMSE:", rmse)

# Obtain metrics of the normal (i.e original) dataset
features = data.drop(target_variable, axis=1)
target = data[target_variable]
rf_normal = RandomForestRegressor(random_state=42)
# Perform k-fold cross-validation (k=5)
cv_predictions = cross_val_predict(rf_normal, features, target, cv=5)
rmse = np.sqrt(mean_squared_error(target, cv_predictions))
print("Normal RMSE:", rmse)
```